# Ensemble of Learning Project Productivity in Software Effort Based on Use Case Points

Mohammad Azzeh
Department of Software Engineering
Applied Science Private University
Amman, Jordan
m.y.azzeh@asu.edu.jo

Ali Bou Nassif
Department of Electrical and Computer Engineering
University of Sharjah,
Sharjah, UAE
anassif@sharjah.ac.ae

Shadi Banitaan
Department of Mathematics, Computer Science and Software Engineering
University of Detroit Mercy, USA
banitash@udmercy.edu

Cuauhtémoc López-Martín
Department of Information Systems
Universidad de Guadalajara
México
cuauhtemoc@cucea.udg.mx

*Abstract*— **It is well recognized that the project productivity is a key driver in estimating software project effort from Use Case Point size metric at early software development stages. Although, there are few proposed models for predicting productivity, there is no consistent conclusion regarding which model is the superior. Therefore, instead of building a new productivity prediction model, this paper presents a new ensemble construction mechanism applied for software project productivity prediction. Ensemble is an effective technique when performance of base models is poor. We proposed a weighted mean method to aggregate predicted productivities based on average of errors produced by training model. The obtained results show that the using ensemble is a good alternative approach when accuracies of base models are not consistently accurate over different datasets, and when models behave diversely.**

I. INTRODUCTION

Use Case Point is one of the most investigated sizing techniques in software engineering literature [1][2]. The basic idea of UCP is to convert elements of use case diagram into its corresponding size metrics that help in estimating amount of work needed to accomplish software projects. To achieve that goal, most of practitioners use basic effort equation to convert UCP size into its corresponding effort in terms of hours/UCP as shown in Equation 1. The procedure of computing UCP is straightforward as described by Karner [3], but the problem is how to find the proper productivity value that effectively lead to accurate effort estimate. The project productivity is defined as ratio between the amount of cost and labor (effort) to the size (UCP) of software project [4][5]. Various authors proposed different approaches and models to predict productivity at early stages of software development. Karner [3] suggested using 20 hours/UCP as generic productivity value irrespective of type and complexity of the software project. Schneider and Winter [6] defined three levels of productivity based on analyzing environmental factors. They proposed a simple algorithm to determine the correct productivity value among 20, 28, and 36 hours/UCP. The main problem of these two approaches that they do not use learning methods to dynamically predict and adjust productivity based on previous projects. In contrast, Azzeh et al. [2][7] proposed a hybrid model based on support vector machine and radial basis neural networks to learn and adjust productivity from historical data. Likewise, Nassif et al. [1] proposed a fuzzy logic based on pre-defined rules to predict productivity. None of the previous studies attempted to construct an ensemble of machine learning models to learn and adjust productivity before computing effort at early stages of software development.

$$Effort = Productivity \times UCP \qquad (1)$$

Ensemble is a learning method that constructs a combination of prediction models through a particular aggregation mechanism, and then generates final solution by taking a weighted vote of their initial solutions [8][9][10]. The principal idea is that methods can be strengthened and produce better results if they cooperate together as a committee with strong methods [9]. Thus, it is useful for the problem of software effort prediction since each single (aka base) model comes with its own assumption and configuration parameters that make ensemble performs extremely well with some desirable statistical properties [8]. Practically, it is highly preferred that the base models used in constructing ensemble to have different characteristics and behave diversely [10]. Diversity means that each model generates statistically significant different predictions than other participant models. In Ensemble, the solution of new problem can be aggregated from a set of initial solutions by applying either simple statistical methods such as mean, weighted mean, inverse ranked weighted method, or by a more complex machine learning based methods such as Bayesian averaging, Bagging and boosting [9]. Thus, the methods in ensembles can boost each other in which estimate errors can be reducing because each method in the ensemble tries to minimize and patch errors made by other methods. Kocaguneli et al. [9] distinguish between two main categories of prediction methods: learner method and solo method. Learner is a single method without supplement of pre or post processing stages. Solo method is a method supplied with pre-processing stage such as normalization and/or feature selection. Accordingly, the term mutli-methods is used to indicate a collection of two or more solo methods. Since learners are not usually supplied with pre or post processing stages, using them are considered copy of one another and have the same bias. Thus, the solo methods are only used to construct ensemble methods because they present different biases and assumptions. All prediction models used in this paper have different assumptions and biases.

In this paper we used seven prediction models which are previously validated and recognized for the problem of effort estimation. These models are used to predict productivity from environmental factors that are available with UCP method. Previous studies have demonstrated that the environmental factors contribute efficiently in predicting productivity because they are originally proposed to reflect the experience and capability of team on the needed effort [6]. We also proposed a new algorithm to construct ensemble among seven models and aggregate predicted productivities. The final predicted productivity obtained by the proposed ensemble model is then used in equation 1 to predict project effort.

## II. RELATED WORK

UCP sizing technique has been studied intensively in literature in the last two decades [11]. The main concern is how to convert the obtained UCP size value into corresponding project effort. Several authors proposed regression models to build relationship between effort and UCP without the need for productivity [7][12]. These models are subject to erroneous and their conclusions cannot be generalized because data collected in these studies are very few. Other authors used productivity as key driver in predicting effort from UCP [1][2]. They proposed various approaches to identify productivity from either UCP size metrics or adjustment factors. In this regard, Karner [3] proposed using 20 hours/UCP as fixed productivity value for all projects irrespective of type and complexity of the software project. Schneider and Winter [6] defined three possible values of productivity based on analyzing environmental factors. They proposed a simple algorithm to find the correct productivity value among 20, 28, and 36 hours/UCP. This algorithm has been examined on large scale datasets and showed worse performance in comparison to regression models. Azzeh et al. [2] proposed different models to learn and adjust productivity from historical data. Similarly, Nassif et al. [1] proposed a fuzzy logic based on pre-defined rules to predict productivity.

The topic of ensemble learning in UCP effort estimation has not been previously examined. Ensemble learning in software effort estimation has been extensively studied in prior studies [8][9][10]. These studies focused the light on various aspects of effort estimation issues such as diversity among base models, ranking of models inside ensembles, aggregation techniques, models selection and weighting. There is no agreement among researchers on the superiority of ensemble in software effort estimation, but one can consider it as a good alternative solution when various base models perform diversely but not accurately enough.

Kocaguneli et al. [9] studied the importance of ranking stability and ensemble methods over 90 solo methods and 20 datasets. The results obtained concluded that the ensemble methods are consistently superior, trustworthy and have smaller error rate. Similarly, both Pahariya et al. [13] also reported improvements over solo-methods. Other studies came to report different conclusions in which that ensemble methods are not quite better than single methods in terms of accuracy and statistical difference [14][15]. Kocaguneli et al. [8] failed to improve the predictive performance of ensemble methods under different scenarios. They used various multi methods combined from 14 different effort estimation methods, applied to two datasets. This study is a replication to Khoshgoftaar et al. [15] study but in the area of software effort estimation. Vinaykumar et al. [14] investigated two kinds of ensembles combined from various learners, but the obtained results were not generally successful. Azzeh e al. [10] investigated various ensembles of adjustment methods in analogy-based estimation. They figured out the importance of ensemble in adjusting predictions.

## III. EVALUATION MEASURES

The choice of evaluation measures is a critical issue in software effort estimation because some evaluation measures are considered biased toward overestimation or underestimation [16][17][18]. In this paper we use three trustworthy measures that have been used in previous studies. The first measure is mean of absolute errors (MAE) which measures average of absolute errors in the prediction model. The remaining measures are Mean Balanced Relative Error (MBRE) and Mean Balanced Inverse Relative Error (MIBRE).

$$MAE = \frac{\sum_i^n |e_i - \hat{e}_i|}{n} \quad (2)$$

$$MBRE = \frac{1}{n}\sum_1^n \frac{AE_i}{min(e_i, \hat{e}_i)} \quad (3)$$

$$MIBRE = \frac{1}{n}\sum_1^n \frac{AE_i}{max(e_i, \hat{e}_i)} \quad (3)$$

Where $e_i$ and $\hat{e}_i$ are the actual and estimated effort of a project.

## IV. THE PROPOSED MODEL

The goal of this study is to examine the efficiency of ensemble learning for predicting project productivity from environmental factors, and hence improve accuracy of effort estimation based on UCP. The UCP method contains two types of adjustment factors, technical and environmental. However, we have chosen environmental factors because they contribute efficiently to productivity as they measure team experience, capabilities and familiarity with project type. Moreover, the way to evaluate these factors are straightforward and do not need professional experts. The eight environmental factors will be used as input variables for project productivity prediction model. Therefore, we identified seven prediction methods that have been used previously in area of software estimation. These models are Regression Tree (RT), Support Vector Regression (SVR), Multiple Linear Regression (MLR), Multi-layer Perceptron neural network (MLP) Radial Basis neural network (RFB), Stepwise Regression (SR) and Fuzzy model. As mentioned earlier, it is preferred that all models behave diversely in order to achieve better accuracy than single base model. But, in fact this is not always a practical option especially when they are not accurate. Therefore, we believe there is a room for improvement even when models do not behave diversely [10]. The Ensemble model can be constructed from multiple models by either: (1) changing data representation such as feature selection and selecting training sets, or (2) by applying architectural methodologies such as Bagging, Boosting and Stacking. In bagging, all ensemble methods are independently applied to different training sets that are selected via bootstrap sampling [9], whereas in Boosting the solo methods are used in a sequence to boost each other for each instance [9]. Both Bagging and Boosting are applied for the same model but with different training data each time, but Stacking can be applied to different models.

In this paper we used bagging for learning the error measures locally and stacking ensemble with weighted mean aggregation to test projects. The process of constructing ensemble in our study is described by the following steps.

**Step 1**: Constructing base models and Learning aggregation weights based error measures.

The goal of this step is to calculate weight of each participated model based on its errors that is measured by *MAE*, *MBRE* and *MIBRE*. First, the training data which contains eight environmental factors, and historical productivity as output are entered into the seven employed methods to construct various local base models using bagging algorithm. In other words, the training data is divided into many subsets of training and testing data to construct local models that help in obtaining *MAE*, *MBRE*, *MIBRE* of local training. Then the evaluation errors of all models are normalized individually (i.e. each measure solely) using min-max approach to have the same influence. To reflect the normalized error values as weight for each model, we used sigmoid function to map them to a discounting factor as shown in equations 2, 3 and 4, where $\overline{MAE}$, $\overline{MBRE}$, and $\overline{MIBRE}$ are the mean of the normalized *MAE*, *MBRE* and *MIBRE* respectively. α is a scaling constant to make weight is close to 1 when any normalized error≃0. In this study we use α=15. The hypothesis of our weighting function is that the model with small error rate is given greater weight than other poor models as shown in Figure 1. In other words, if the normalized *MAE* of a particular model is very low then large weight is given to that model. In contrast, if the normalized *MAE* is high then low weight is given to that model. Finally, since there are three evaluation measures, we treat each measure individually in the same manner, then we take average of these weights to produce the corresponding weight of the model as shown in Equation 7.

$$w^i_{MAE} = \frac{1}{1 + e^{\alpha(MAE_i - \overline{MAE})}} \quad (4)$$

$$w^i_{MBRE} = \frac{1}{1 + e^{\alpha(MBRE_i - \overline{MBRE})}} \quad (5)$$

$$w^i_{MIBRE} = \frac{1}{1 + e^{\alpha(MIBRE_i - \overline{MIBRE})}} \quad (6)$$

$$w_i = \frac{w^i_{MAE} + w^i_{MBRE} + w^i_{MIBRE}}{3} \quad (7)$$

Where $w^i_{MAE}$, $w^i_{MBRE}$ and $w^i_{MIBRE}$ are weight obtained from normalized error measure *MAE*, *MBRE* and *MIBRE* for model *i*.

**Step 2**: productivity prediction

In this step, the test data is entered into the constructed seven models to produce seven productivity values. These productivity values are aggregated using weighted mean equation as shown in Equation 8. The weights used here are those obtained from step 1.

$$productivity_j = \frac{\sum_{i=1}^{7} productivity_i \times w_i}{\sum_{i=1}^{7} w_i} \quad (8)$$

Where $productivity_j$ is the aggregated productivity of the test project. $productivity_i$ is the productivity of test project obtained by model *i*.

**Step 3**: the predicted productivity from step 2 and the available UCP of test project are multiplied by each other as shown in equation 1 to predict final effort of the test project.

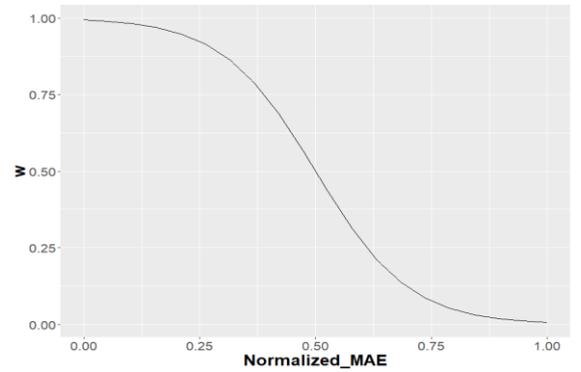

Figure 1. Sigmoid discounting function with $\alpha = 15$

V. DATASETS

Two datasets have been collected and used for estimating effort based on Use Case Points sizing technique (hereafter both datasets are called DS1 and DS2). DS1 contains information about software projects collected from IT company. These projects are collected from various industrial and governmental sectors, using 2-tier and 3-tier software application architecture. DS2 contains information about fourth year and master student projects, developed for educational purposes. These projects were developed using object-oriented analysis, design and programming methodologies. Tables 1 and 2 describe the statistical properties of both datasets. From both table we can notice that student projects are more productive than industrial projects as they need less time to finish one UCP. The productivity variable in both datasets are normal as confirmed by Skewness measure. The kurtosis of productivity variable suggests that it is normally distributed but with shorter tail and wide peak.

TABLE 1 Descriptive statistics of DS1 dataset

| Variable | Mean | StDev | Skewness | Kurtosis |
|---|---|---|---|---|
| UCP | 739.3 | 1563.9 | 3.0 | 11.7 |
| Effort | 20573.5 | 47326.9 | 3.2 | 12.4 |
| productivity | 24.1 | 5.1 | 0.0 | 2.2 |

TABLE 2 Descriptive statistics of DS2 dataset

| Variable | Mean | StDev | Skewness | Kurtosis |
|---|---|---|---|---|
| UCP | 82.6 | 20.7 | 0.8 | 4.1 |
| Effort | 1672.4 | 414.3 | -0.1 | 2.2 |
| productivity | 20.8 | 4.8 | 0.2 | 2.7 |

## VI. RESULTS

Constructing ensembles from many prediction models is not trivial task, and considered time consuming. The typical way to construct ensemble is either to reduce number of models where only superior models are ranked and used only, or using all models but with different impacts and influences as we did in this study. The procedure of constructing ensemble has been explained in Section 4. All models have been validated using Leave one out cross validation in which one project is used as test and the remaining projects are used as training cases [19][20]. This procedure continues until all projects act as test cases. For each model, we record *MAE*, *MBRE* and *MIBRE*. Tables 3 and 4 show the empirical validation results over DS1 and DS2. It is clear from the tables that using ensemble of models with different weights based on local training errors is more superior than using base models. In each run of the empirical validation, the models with minimum error rates are given higher weight than others. This weight frequently changes from one test case to another based on the structure of dataset. Changing the weights for each test project allows us to dynamically adjust aggregation process based on the training data structure. However, the difference between ensemble and other base models is clear over DS1, but it is not much clear over DS2. The main reason for that, DS2 contains projects that were developed by students at university so the quality of data collection process is not yet mature as for industrial projects which usually performed by experts. If we look closer at the results we can notice that there is no consistency among evaluation measures for the superiority. In fact, this is big challenge in software effort estimation area, in spite of many studies attempt to propose evaluation framework to consistently evaluate prediction models [21]. The results of MAE in both datasets confirm that the ensemble surpasses other based models. This is true as for other evaluation measure but with little effect.

We also ran Wilcoxon test to statistically measure the significant difference between ensemble model and other base models over two datasets, but unfortunately, we did not find any significant differences. We also did not find significant difference between base models. This might explain why differences between models in terms of *MBRE* and *MIBRE* are not noticeable as in *MAE*. However, we can still encourage with the results even though the base models are not diverse enough.

TABLE 3 Accuracy results over DS1

|  | **MAE** | **MBRE** | **MIBRE** |
|---|---|---|---|
| **Ensemble** | 1744.271 | 0.137992 | 0.103015 |
| **MLR** | 2035.541 | 0.155744 | 0.122318 |
| **Fuzzy** | 2000.32 | 0.152436 | 0.121872 |
| **SVR** | 1992.856 | 0.149145 | 0.118567 |
| **SR** | 2136.459 | 0.158958 | 0.124325 |
| **RT** | 3406.502 | 0.201927 | 0.153181 |
| **MLP** | 4169.979 | 0.238815 | 0.169546 |
| **RBF** | 2188.861 | 0.148738 | 0.115815 |

TABLE 4 Accuracy results over DS2

|  | **MAE** | **MBRE** | **MIBRE** |
|---|---|---|---|
| **Ensemble** | 262.759 | 0.177128 | 0.130254 |
| **MLR** | 270.6724 | 0.193445 | 0.143428 |
| **Fuzzy** | 344.1473 | 0.270714 | 0.184855 |
| **SVR** | 297.2925 | 0.215937 | 0.157825 |
| **SR** | 287.284 | 0.201706 | 0.149541 |
| **RT** | 274.4744 | 0.198441 | 0.146968 |
| **MLP** | 385.7305 | 0.275806 | 0.192511 |
| **RBF** | 275.1779 | 0.199447 | 0.1484 |

Figures 2 and 3 show the interval plots for all models over both datasets. The confidence bar length represents the range of values that is likely to include the population mean while the circle is the mean of confident population. From Figure 2, we can notice that both the constructed ensemble and SR have relatively the same mean but the length of interval for ensemble model is shorter which confirms that ensemble generates better predictions. In Figure 2, all models have substantially similar bar length but with different absolute error means. However, the interval bars in both figures are overlapping which might confirm that all models are substantially not different from each other.

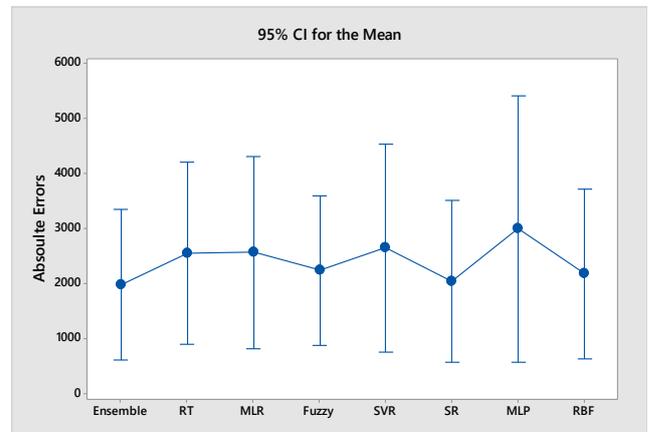

Figure 2. confidence error bar plot for all models over DS1

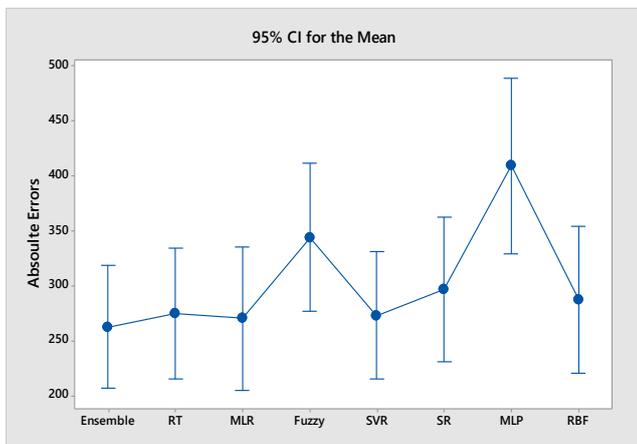

Figure 3. confidence error bar plot for all models over DS2

## VII. Conclusions

This paper presents a new ensemble learning model to predict productivity from seven well known base models. These models have been selected because they behave diversely, and they use different procedure to produce predictions. The ensemble is constructed and aggregated using weighted mean, where weights are found based on analyzing training data errors throughout bagging procedure. The results obtained are accurate in comparison with participated base models. Surprisingly, we have noticed that even though that base models are not substantially diverse, the ensemble works well when predictions are aggregated using weighted mean. Further studies are needed to compare our approach to other ways of ensemble construction.

### Acknowledgment

Mohammad Azzeh is grateful to the Applied Science Private University, Amman, Jordan, for the financial support granted to cover the publication fee of this research.

Ali Bou Nassif would like to thank the University of Sharjah for the continuous research support.